\documentclass[11pt]{article}
\usepackage[margin=1in]{geometry}
\usepackage{amsmath,amssymb}
\usepackage{booktabs}
\usepackage{graphicx}
\usepackage{caption}
\usepackage[hidelinks]{hyperref}
\usepackage{xcolor}
\usepackage{natbib}
\usepackage{titlesec}
\titlespacing*{\section}{0pt}{1.2em}{0.5em}

\newcommand{\code}[1]{\texttt{\small #1}}

\title{\bf The Filter Metric is Safety-Critical:\\
Phantom Advantages in Group-Relative RL under Shaped Rewards}
\author{Juntao Yu\thanks{Fudan University, Shanghai, China. ORCID:
\href{https://orcid.org/0009-0002-6878-7930}{0009-0002-6878-7930}.
Correspondence: \texttt{2750167653@qq.com}. Companion artifact:
\href{https://github.com/yuyu0529nya/rl-filter-metric-study}{rl-filter-metric-study}; public core
recipe: \href{https://github.com/verl-project/verl-recipe/pull/125}{verl-recipe PR \#125}.}}
\date{August 10, 2026}

\begin{document}
\maketitle

\begin{abstract}
Group-relative policy optimization (GRPO and descendants) can discard ``no-contrast'' rollout
groups through dynamic sampling, while practical implementations expose a \emph{configurable
filter metric}. We identify and quantify a metric--predicate mismatch under composite shaped
rewards. When the filter follows the shaped training score rather than the task outcome, all-fail
groups retain nonzero within-group spread and pass the predicate; standard-deviation normalization
then promotes shaping differences among failures to full-size \emph{phantom advantages}. In a
controlled three-arm comparison on GSM8K (Qwen2.5-1.5B, LoRA), no filtering and shaped-score
filtering end at EM $0.080 \pm 0.112$ and $0.040 \pm 0.008$, whereas binary-outcome filtering
holds $0.754 \pm 0.005$ across four runs per arm (three default-seed reruns and one seed-123 run;
mean $\pm$ sample SD). On verl's native
\code{recipe/dapo} trainer, holding model, data, reward and trainer fixed and changing only the
metric, the score arm requires no batch refill in any of 40 observed steps and ends at EM $0.160$;
the accuracy arm refills in 29/40 steps and ends at $0.763$. Both use the same custom shaped-reward
hook; neither modifies trainer or filtering code. Prior work has established shaping-induced
amplification and all-fail filtering; our contribution isolates the metric--predicate semantic
mismatch and directly instruments native deletion/refill telemetry. Across the tested positive
coefficients $\lambda \in \{0.1,0.3,0.5\}$, the unsafe arms collapse, while reversed-length and
format probes do not collapse within 40 steps. Exploratory one-run cells reproduce the failure on
MATH at 1.5B/7B and under GSPO; disabling standard-deviation normalization (Dr.~GRPO) avoids the
observed collapse at both scales. The practical implication is simple: filtering under a composite
reward should use a task-outcome signal whose semantics are independent of the shaping term.
\end{abstract}

\section{Introduction}

Critic-free group-relative estimators normalize rewards within a group of rollouts sharing a
prompt, $A_i = (r_i - \mathrm{mean}(r)) / (\mathrm{std}(r) + \epsilon)$. With a binary reward, an
all-same group has $\mathrm{std} = 0$ and advantage exactly $0$; DAPO's dynamic sampling
\citep{yu2025dapo} discards such groups, and its published criterion is accuracy-based. The verl
implementation \citep{sheng2024hybridflow,verldapo2026} generalizes this to ``drop groups whose
configurable \code{filter\_groups.metric} is all-same.'' The published and documented DAPO recipes
use \code{metric=acc}. Documented alternatives include \code{score} and \code{seq\_reward}; their
semantics depend on the reward manager. In our hook, \code{score} stores the composite shaped score.

This paper measures what that field is worth. Our contributions:
\begin{enumerate}\itemsep2pt
\item a controlled measurement of the metric--predicate mismatch under a shaped reward
      ($0.040$ versus $0.754$ EM for shaped-score versus outcome filtering);
\item a native-trainer control-flow measurement: switching only the metric changes whether the
      physical refill loop is exercised (0/40 versus 29/40 steps) and changes EM from $0.160$ to
      $0.763$;
\item boundary probes across coefficient, shaping direction, model/task scale and estimator,
      with replication levels stated explicitly; and
\item a conditional negative result: re-zeroing binary-reward dead groups is gradient-redundant,
      whereas physical deletion and refill can still change the sampled prompt distribution.
\end{enumerate}

This is explicitly \textbf{not} a defect report against DAPO: its published and official recipe
uses the outcome-based arm of our comparison. We study a documented configuration surface and the
semantic mismatch that appears when its metric is repurposed for a composite score.

\section{Related work}

\paragraph{Group-relative RL.} GRPO \citep{shao2024deepseekmath} removes the critic by normalizing
within groups and underpins R1-style reasoning training \citep{deepseek2025r1}. DAPO
\citep{yu2025dapo} adds dynamic sampling: over-sample prompts and filter out those whose rollouts
are all-correct or all-incorrect. GSPO \citep{zheng2025gspo} moves the importance ratio and
clipping to the sequence level. Dr.~GRPO \citep{liu2025drgrpo} identifies an optimization bias in
GRPO that inflates response length, particularly for incorrect outputs. Dr.~GRPO also enters our
results as a mechanism intervention because it removes standard-deviation normalization
(\S\ref{sec:algo}).

\paragraph{Shaping under group normalization.} Several studies already establish the broad
amplification mechanism. Pref-GRPO observes that normalization can turn minimal pointwise reward
gaps into illusory advantages \citep{wang2025prefgrpo}. GR$^3$ analyzes optimization shortcuts
created by additive length shaping \citep{li2026gr3}. ACOER shows that continuous length penalties
on incorrect answers can collapse GRPO and proposes correct-only shaping \citep{lee2026acoer}.
Most directly, the Dark Room study formalizes coefficient cancellation in all-fail groups and
shows that removing standard-deviation normalization or dropping all-fail groups can prevent the
collapse \citep{wang2026darkroom}. We therefore do not claim to discover shaping-induced
amplification, correct-only gating, or all-fail filtering. Our narrower question is
implementation-level: when a DAPO implementation exposes the group-filter metric as a configurable
field, can a composite shaped score change the predicate's semantics and suppress the native
deletion-and-refill control flow? We answer this with a metric-only comparison and refill telemetry.

\paragraph{Reward hacking and Goodhart's law.} Optimizing an imperfect proxy can decrease true
performance \citep{skalse2022rewardhacking, gao2022overoptimization}. That literature largely
studies learned reward models drifting from human preference; the failure here is narrower---a
hand-written shaping term interacting with a group-normalizing estimator. The implementation-level
twist is that outcome-degenerate groups can pass silently when the configured metric no longer
represents the task outcome. Our finite-horizon probes further suggest asymmetric damage: shortcuts
that erode task completion are more destructive than the completion-preserving alternatives tested.

\paragraph{Process rewards.} Step-level supervision can improve reasoning over outcome-only
supervision \citep{lightman2023verify}. Dense feedback is especially attractive when outcomes are
sparse---the regime targeted by our analysis.

\section{The phantom advantage}

We use a shaped reward with a tunable phantom strength $\lambda$ (length in characters):
\[
r = \mathrm{outcome}(0/1) - \lambda \cdot \min(\mathrm{len}/512,\ 1).
\]
Among failures the shortest wrong answer scores best (``give up fast''). The probe is distilled
from a real incident: a search agent whose answer length collapsed from $\sim$24 to $\sim$7
characters under a hand-tuned shaped reward. Under this reward an all-fail group has
$\mathrm{std} > 0$, survives a shaped-score filter, and its normalized advantages point toward
shorter failures. It carries no binary outcome contrast and here favors shorter failures. This is an instance of the amplification
mechanism established in prior work \citep{lee2026acoer,wang2026darkroom}; here it serves as a
controlled probe of the filter metric and its downstream control flow.

\section{Controlled design}

The three arms differ only in which groups receive zero advantage. We implement the gate with a
wrapper around verl's advantage computation:

\begin{center}\small
\begin{tabular}{lll}
\toprule
arm & zeroes (advantage $\to 0$) & corresponds to \\
\midrule
no-filter & nothing & plain GRPO \\
shaped-score & groups with shaped-reward $\mathrm{std} \approx 0$ & \code{metric = score / seq\_reward} \\
binary-outcome & groups with all-same binary outcome & \code{metric = acc} (official DAPO scripts) \\
\bottomrule
\end{tabular}
\end{center}

Two design decisions matter. First, we \textbf{zero advantages, not the response mask}. For rows
whose advantage is already zero, mask-dropping only changes the token-mean loss scale in our
implementation. Our controlled A/B finds no consistent endpoint effect (\S\ref{sec:neg});
advantage-zeroing keeps the denominator fixed. Second, the gate reads the \emph{raw outcome}, not
the shaped scalar, so ``barely failed'' is never conflated with ``passed''.

\paragraph{Setup.} We use Qwen2.5-1.5B/7B-Instruct \citep{yang2024qwen25}, LoRA $r{=}32$
\citep{hu2021lora}, five rollouts per prompt, 32 prompts per batch, lr $10^{-4}$ and 40 steps.
Held-out validation uses all 1,319 GSM8K items \citep{cobbe2021gsm8k}. For MATH
\citep{hendrycks2021math}, we sample 500 items at seed 42; filtering two overlength examples gives
effective $n{=}498$.
Hardware comprises two RTX 5090 32\,GB GPUs and one RTX PRO 6000 96\,GB GPU.

\section{Experiment 1: core GSM8K-1.5B contrast}

\begin{figure}[t]\centering
\includegraphics[width=\linewidth]{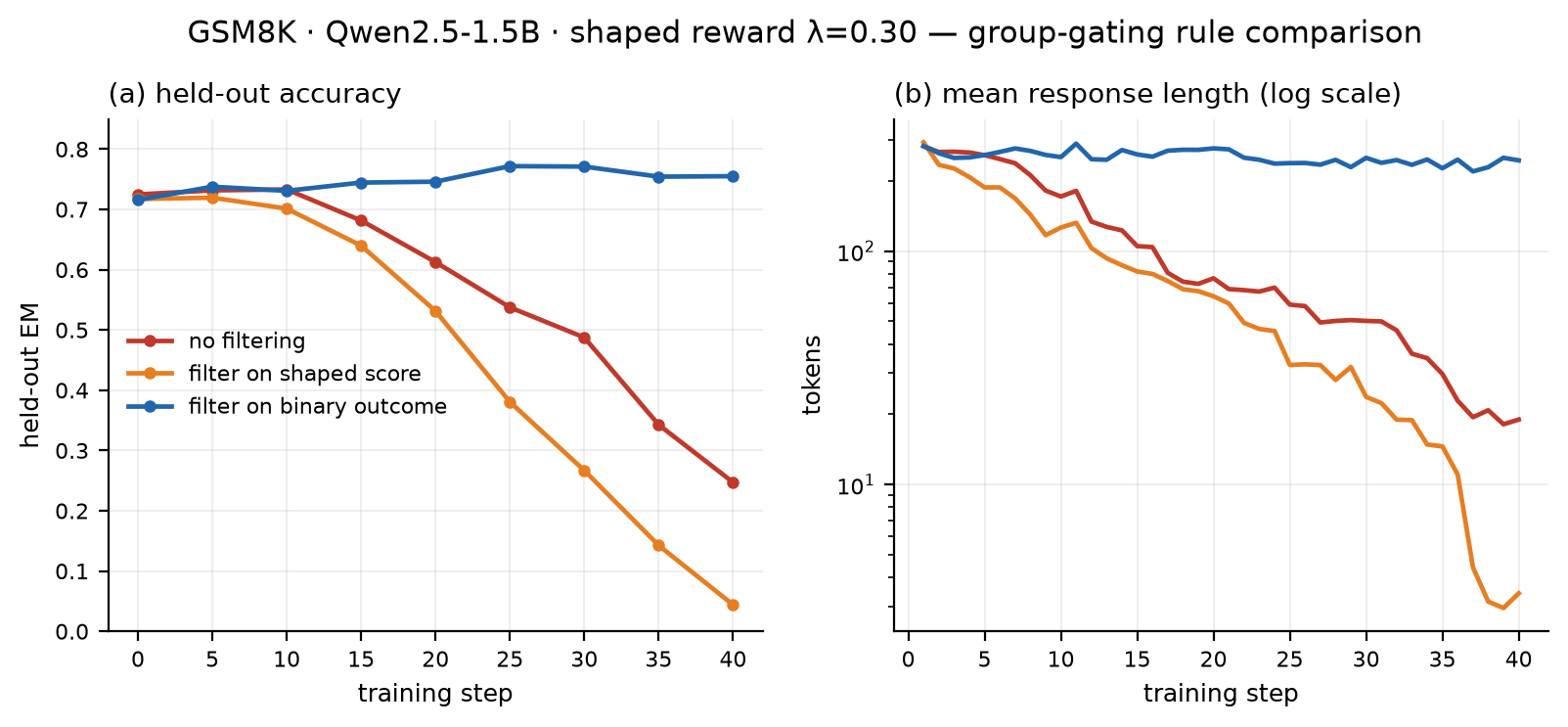}
\caption{Same model, data and hyperparameters; only the group-gating rule differs. Right: the
mechanism---both unsafe arms are driven to very short final responses (roughly 3--18 tokens).}
\end{figure}

\begin{center}\small
\begin{tabular}{lccccc}
\toprule
val EM @ step 40 & rep 1 & rep 2 & rep 3 & seed 123 & pooled \\
\midrule
no-filter & 0.247 & 0.038 & 0.022 & 0.014 & $0.080 \pm 0.112$ \\
shaped-score & 0.044 & 0.031 & 0.037 & 0.049 & $0.040 \pm 0.008$ \\
\textbf{binary-outcome} & 0.755 & 0.758 & 0.747 & 0.754 & $\mathbf{0.754 \pm 0.005}$ \\
\bottomrule
\end{tabular}
\end{center}

All arms start from val@0 $\approx 0.715$. Runs 1--3 use the framework default seed and differ
through rollout/scheduling nondeterminism; the fourth run sets
\code{rollout.seed = data.seed = 123}. We therefore report four runs, not four distinct seeds.
Telemetry: while training is healthy the shaped-score
filter drops only $\sim$2/32 degenerate groups per batch---the phantom groups sail through---while
the outcome filter drops 16--24/32.

\section{Experiments 2 and 3: tested doses and shaping direction}

\begin{figure}[t]\centering
\includegraphics[width=0.62\linewidth]{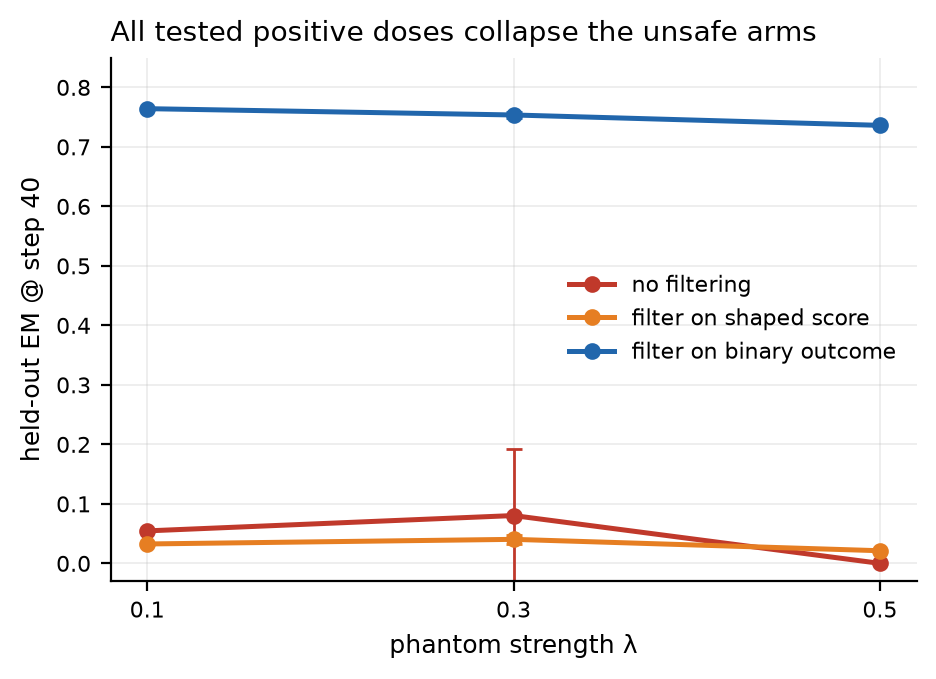}
\caption{Endpoint EM versus phantom strength $\lambda$. Error bars on the $\lambda{=}0.30$ anchor
are sample standard deviations across four runs per arm.}
\end{figure}

\begin{center}\small
\begin{tabular}{lccc}
\toprule
val@40 & $\lambda{=}0.10$ & $\lambda{=}0.30$ & $\lambda{=}0.50$ \\
\midrule
no-filter & 0.055 & $0.080 \pm 0.112$ & \textbf{0.000} \\
shaped-score & 0.033 & $0.040 \pm 0.008$ & 0.021 \\
binary-outcome & 0.764 & $0.754 \pm 0.005$ & 0.736 \\
\bottomrule
\end{tabular}
\end{center}

No positive coefficient in the tested grid avoids collapse in the unsafe arms; larger $\lambda$
accelerates it. At $\lambda{=}0.50$ plain GRPO reaches EM $0.000$ with mean response length $1.0$
token, consistent with exploiting the shortest-failure shortcut. The outcome filter's mild slope
($0.764 \to 0.736$) is consistent with the residual length penalty applied to correct answers
inside live groups. This three-point sweep does not establish a universal threshold or rule out a
smaller finite-horizon-safe coefficient.

In a one-run probe, reversing the phantom so the \emph{longest} wrong answer scores best
($\lambda{=}0.30$) yields endpoints $0.736 / 0.746 / 0.754$ without collapse. This probe suggests
greater risk when the shortcut destroys task
structure: truncation deletes the answer, whereas padding leaves it intact and, in mixed groups,
the $+1.0$ for a correct answer dwarfs the shaping term.

\paragraph{A second shaping family: format.} Length is not the only way to shape a reward, so we
repeated the three arms with a \emph{format} phantom instead: among failures, a wrong answer
carrying the expected answer marker (\code{\#\#\#\#}~$\langle$number$\rangle$) scores $0$ while one
without it scores $-\lambda$. This is not a contrived probe---format rewards are standard in RLVR
pipelines. No collapse is observed in these one-run cells (endpoints $0.745 / 0.736 / 0.769$ for
no-filter, shaped-score and binary-outcome respectively).

Taken together, the three probes sharpen the asymmetry from a statement about \emph{direction}
into one about \emph{structure}:

\begin{center}\small
\begin{tabular}{lll}
\toprule
phantom & what the shortcut does & outcome \\
\midrule
length, short-favoring & truncates the answer away & very short outputs (as low as 1 token) \\
length, long-favoring & pads an intact answer & no collapse observed \\
format & produces well-formed wrong answers & no collapse observed \\
\bottomrule
\end{tabular}
\end{center}

These probes support a finite-horizon hypothesis: shaping is most dangerous when its shortcut
directly erodes task completion, whereas a completion-preserving shortcut can be less destructive.
The reverse-length and format cells are $n{=}1$, so they are boundary probes rather than evidence
that either shaping family is generally safe.

\section{Experiment 4: scale and difficulty}

\begin{figure}[t]\centering
\includegraphics[width=\linewidth]{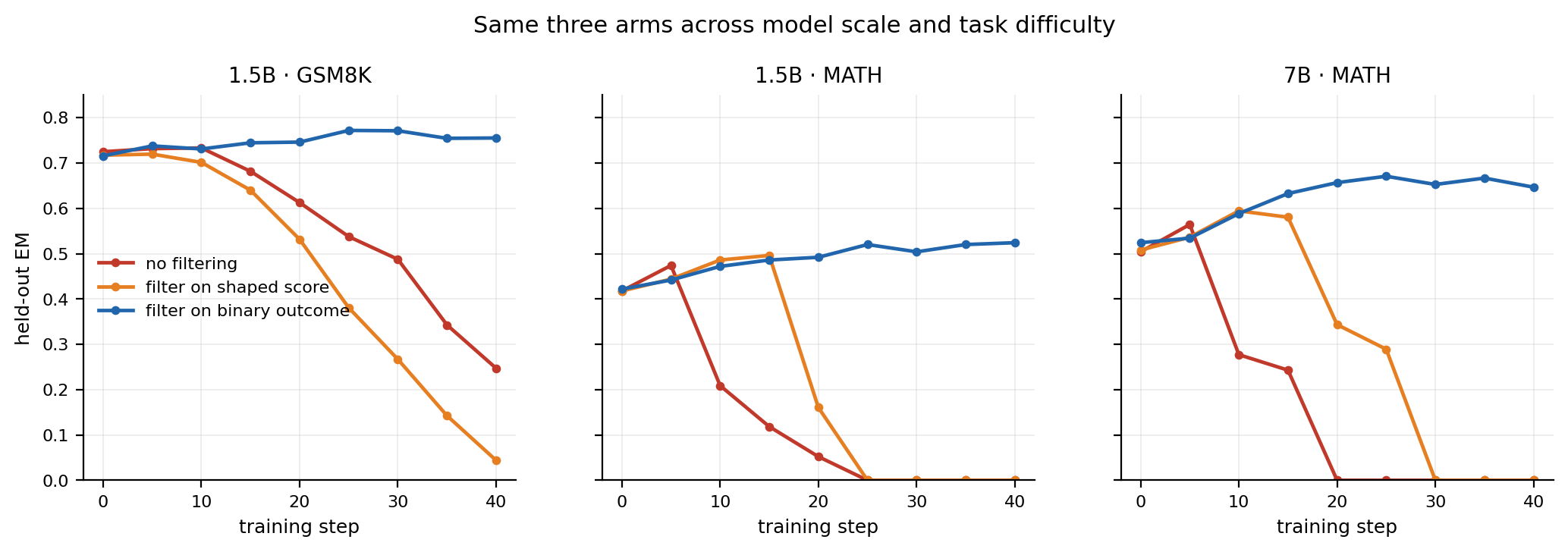}
\caption{The same three arms across model scale and task difficulty. On MATH, unsafe arms at both
scales reach EM $0$ within 20--30 steps, while outcome filtering keeps learning.}
\end{figure}

\begin{center}\small
\begin{tabular}{lccc}
\toprule
val EM @ 40, $\lambda{=}0.30$ & no-filter & shaped-score & binary-outcome \\
\midrule
1.5B $\cdot$ GSM8K (anchor, $n{=}4$) & $0.080 \pm 0.112$ & $0.040 \pm 0.008$ & $\mathbf{0.754 \pm 0.005}$ \\
1.5B $\cdot$ MATH & 0.000 & 0.000 & $\mathbf{0.524}$ ($+10.2$ pp) \\
7B $\cdot$ MATH & 0.000 & 0.000 & $\mathbf{0.647}$ ($+12.3$ pp) \\
7B $\cdot$ GSM8K & \multicolumn{3}{c}{\emph{no dose---see below}} \\
\bottomrule
\end{tabular}
\end{center}

\textbf{Not a small-model artifact.} On MATH, unsafe arms at both 1.5B and 7B reach EM $0$ by step
20--30, while the outcome arms learn, gaining $+10.2$ and $+12.3$ percentage points respectively.

\textbf{Dose is set by task difficulty relative to the model.} A canary run on 7B$\cdot$GSM8K
measures val@0 $= 0.913$ and \textbf{0--1 all-fail groups out of 32} per batch: the phantom has very
low measured exposure in those sampled batches, so little collapse pressure is expected in that
observed window. This is an exposure measurement, not proof of immunity. The practical concern is
that shaping is often introduced when a task is hard for the current model---the regime with more
all-fail groups.

\textbf{``Removes the corpse, never the poison'' replicates at scale.} At the moment of death the
shaped-score filter reports \code{dead\_fail = 32/32}---every group all-fail with identical
outputs---i.e.\ it reaches 32/32 after collapse has homogenized the batch. Captured once at
each scale.

\section{Confirmation on the official DAPO trainer}

\begin{figure}[t]\centering
\includegraphics[width=\linewidth]{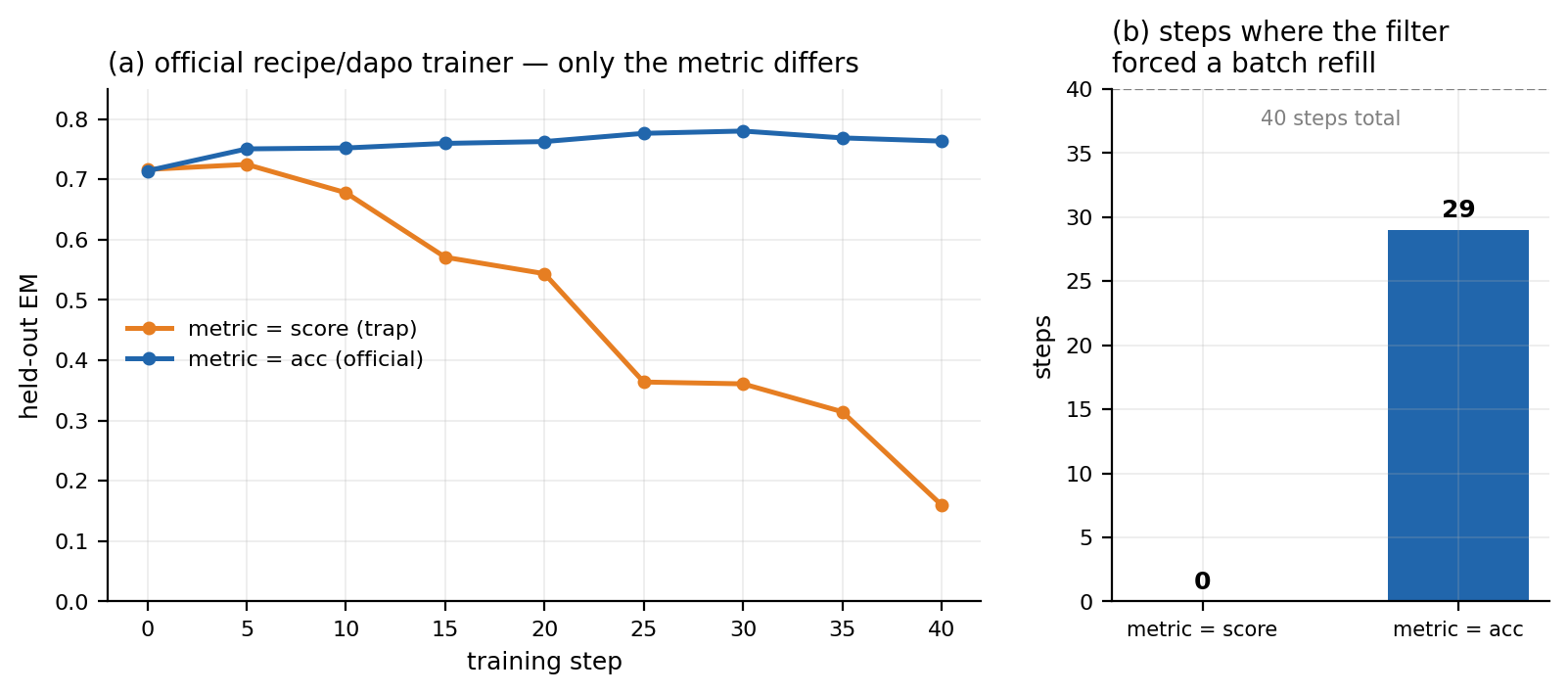}
\caption{Confirmation on verl's official \code{recipe/dapo} trainer with physical group deletion
and batch refilling. Right: one generation batch always suffices for the shaped-score metric,
whereas the accuracy metric exercises the refill path in 29 of 40 steps.}
\end{figure}

The arms above zero advantages to keep the token-mean denominator fixed. To rule out that this
design choice carries the effect, we re-ran the contrast on verl's official \code{recipe/dapo}
trainer---real group deletion plus resampling to refill the batch (\code{gen\_batch\_size}=96,
\code{train\_batch\_size}=32, \code{max\_num\_gen\_batches}=10). Both arms use the same custom
shaped-reward hook, but make no modification to trainer or filter implementation; only
\code{filter\_groups.metric} changes:

\begin{center}\small
\begin{tabular}{lccc}
\toprule
official DAPO, GSM8K-1.5B, $\lambda{=}0.30$ & val@0 $\to$ @40 & mean length, first $\to$ last & refill steps \\
\midrule
\code{metric = score} & $0.716 \to \mathbf{0.160}$ & $265.6 \to 13.7$ tokens & \textbf{0 / 40} \\
\code{metric = acc} & $0.714 \to \mathbf{0.763}$ & $270.8 \to 309.7$ tokens & \textbf{29 / 40} \\
\bottomrule
\end{tabular}
\end{center}

The refill counter is the sharper finding. Under \code{metric=score}, one generation batch is
sufficient in all 40 steps: shaped rewards give nearly every all-fail group nonzero spread, so the
first oversized batch retains enough groups and the refill path is never exercised. This does
\emph{not} mean that the predicate code is disabled or that no group is removed. Under
\code{metric=acc}, 29 of 40 steps need a second generation batch (69 generation batches total,
$1.725\times$ the score arm), and the policy gains $+4.9$ EM while its sibling loses $55.6$. Thus a
valid-looking configuration can retain many outcome-degenerate groups and leave the refill path
inactive without an exception or warning.

\section{Estimator intervention}\label{sec:algo}

If the phantom's lethality comes from group normalization rescaling tiny within-group
differences to full size, then an estimator that does \emph{not} rescale should reduce that
pressure. We test this by re-running the same three arms under two other estimator variants, both
available as native configuration flags in verl (no estimator-code changes): GSPO
(\code{policy\_loss.loss\_mode=gspo} with \code{loss\_agg\_mode=seq-mean-token-mean}) and
Dr.~GRPO (\code{algorithm.norm\_adv\_by\_std\_in\_grpo=false}, i.e.\ $A_i = r_i - \mathrm{mean}(r)$
without the $\mathrm{std}$ divisor).

\begin{center}\small
\begin{tabular}{lccc}
\toprule
val EM @ 40, GSM8K-1.5B, $\lambda{=}0.30$ & GRPO (anchor, $n{=}4$) & GSPO & Dr.~GRPO \\
\midrule
no filtering & $0.080 \pm 0.112$ & 0.030 & \textbf{0.718} \\
filter on shaped score & $0.040 \pm 0.008$ & 0.000 & \textbf{0.747} \\
filter on binary outcome & $\mathbf{0.754 \pm 0.005}$ & 0.734 & 0.738 \\
\midrule
final length (tokens) & 3--18 (unsafe) & 1--2 (unsafe) & 151--202 (all arms) \\
\bottomrule
\end{tabular}
\end{center}

\paragraph{GSPO: the failure is not tied to the importance-ratio granularity.} The collapse and
outcome-filtered recovery recur in these one-run cells. Notably \code{gspo} with shaped-score
filtering is the fastest collapse we observed---EM $0.712 \to 0.153$ by step 5 and $0$ by step 25.
Because this cell is a single run without an aggregation ablation, we do not attribute the speed
difference to GSPO's sequence-level aggregation.

\paragraph{Dr.~GRPO avoids collapse in the tested cell.} All three one-run arms stay healthy and no
length collapse occurs. The arithmetic explains why this intervention should weaken the failure.
Take an all-fail group
whose shaped rewards span roughly $[-0.30, -0.05]$, so $\mathrm{std} \approx 0.09$. With
normalization, $A = (r - \mathrm{mean})/\mathrm{std}$ spans $\approx \pm 1.4$: full magnitude, as
loud as any genuinely informative group. Without it, $A = r - \mathrm{mean}$ spans $\approx \pm
0.12$, while a mixed group (some correct, some not) still spans $\approx \pm 0.7$---roughly six
times louder. Std normalization, in other words, gives every group the same speaking volume
regardless of whether it has anything to say.

\paragraph{The avoidance recurs at 7B on MATH.} A single-scale mechanism claim is weak, so we
repeated the three-arm comparison under Dr.~GRPO at a setting where standard GRPO is destructive.
Under GRPO, both unsafe arms reach
EM $0$ (by step 20 and step 30) with final response lengths of 1--2 tokens. Under Dr.~GRPO, all
three arms not only survive---they learn.

\begin{center}\small
\begin{tabular}{lcc}
\toprule
val EM @ 40, MATH-7B, $\lambda{=}0.30$ & GRPO & Dr.~GRPO \\
\midrule
no filtering & 0.000 & \textbf{0.675} \\
filter on shaped score & 0.000 & \textbf{0.639} \\
filter on binary outcome & 0.647 & 0.665 \\
\midrule
final response length (tokens) & 1--2 (unsafe arms) & 288--333 (all arms) \\
\bottomrule
\end{tabular}
\end{center}

\noindent The two cells that were total losses under GRPO land at $0.675$ and $0.639$, above their
respective Dr.~GRPO starting points of $0.512$ and $0.520$: removing the $1/\mathrm{std}$ factor
turns the trap configurations from catastrophic into non-collapsed learning trajectories. Nothing
else changed---same shaped reward, same $\lambda$, same data, same 40 steps, same machine. Together
with the 1.5B cell, this mechanism intervention implicates standard-deviation amplification as a
causal contributor, but does not establish general Dr.~GRPO immunity.

\paragraph{Two interventions targeting two ingredients.} In our tested cells, collapse is
associated with (i) shaping-induced within-group spread among failures and (ii) std normalization
that rescales that spread. Outcome filtering excludes the affected all-fail groups; dropping std
normalization removes ingredient (ii). Either intervention avoids collapse in these cells.
Dr.~GRPO was
proposed to correct a length bias that makes responses \emph{longer} \citep{liu2025drgrpo}; here
the same normalization change also weakens an explicitly shaped pressure toward shorter failures.

\paragraph{Caveat.} In the GSM8K estimator table, each GSPO and Dr.~GRPO cell is $n{=}1$ and the
GRPO anchor is $n{=}4$. Every MATH-7B cell in the second table is $n{=}1$. What carries the weight is
the repeated pattern: removing amplification coincides with removal of collapse at 1.5B on GSM8K
and at 7B on MATH. We treat this as mechanistic evidence, not a performance ranking or universal
robustness guarantee.

\section{Negative results and limitations}\label{sec:neg}

\paragraph{Re-zeroing binary dead groups is gradient-redundant.} Under binary rewards, an all-same
group already has centered advantage $0$; explicitly setting its advantages to zero again is
algebraically redundant. Mask-dropping those rows without refill additionally changes the
token-mean denominator by $1/\mathrm{live\_frac}$, a loss-scale change that Adam largely absorbs in
our runs. A 20-step, 2-seed $\times$ 2-arm controlled A/B on the airline domain of
$\tau^2$-bench \citep{barres2025tau2} with a 7B policy shows no consistent effect in held-out mean
reward (endpoints $0.5625/0.55$ with the gate off versus $0.4125/0.5375$ with it on).
A single-seed claim that ``gating hurts by $-0.15$'' disappears on the second seed. This result
does not apply to DAPO-style physical deletion and refill, which changes prompt sampling and
gradient variance even when deleted groups would have zero gradient.

\paragraph{Robustness to verifier noise.} Real binary signals come from imperfect verifiers. We
flip the outcome with probability $p$ (deterministic per-sample hash; measured flip rate $0.145$ at
$p{=}0.15$), letting the noise flow through the entire channel---shaped score, filter signal and
validation---and de-noise validation analytically via $\mathrm{true} = (\mathrm{obs} - p)/(1 - 2p)$.
Outcome filtering on GSM8K-1.5B, $\lambda{=}0.30$: at $p{=}0.05$ the de-noised endpoint is $0.745$
and at $p{=}0.15$ it is $0.756$, against a clean anchor of $0.754 \pm 0.005$. Mechanism telemetry
shows why this is nontrivial: noise breaks up all-same groups, so the filter's final-step coverage
degrades sharply (live fraction $0.22 \to 0.50 \to 0.69$), i.e.\ the gate ends up dropping under a third of
groups at $p{=}0.15$, about 40\% of the clean drop fraction. Yet performance holds in these
one-run noise cells, suggesting margin in this setting.

\begin{figure}[t]\centering
\includegraphics[width=0.62\linewidth]{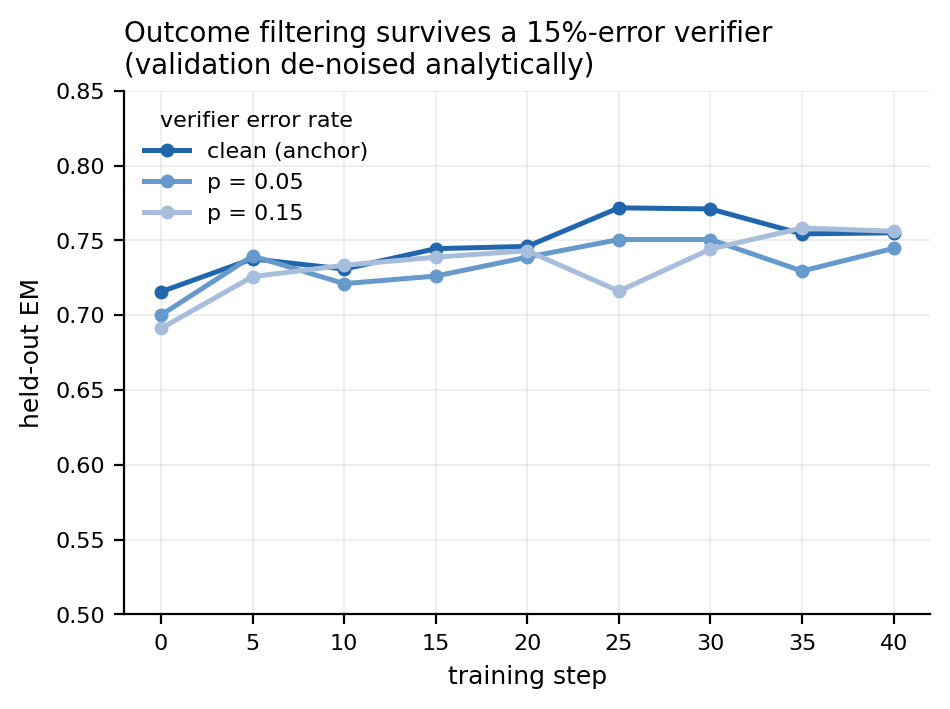}
\caption{De-noised held-out accuracy under a faulty verifier.}
\end{figure}

\paragraph{Limitations.} Only the 1.5B$\cdot$GSM8K anchor has four runs, and just one uses an
explicitly different seed; off-anchor, estimator, shaping-direction and official-DAPO
cells are $n{=}1$. We test one model family, two tasks, LoRA fine-tuning, three estimator variants
and two hand-written shaping families over 40 steps. The reverse-length and format probes show only
that collapse was not observed in that horizon. The 7B$\cdot$GSM8K cell is a measured absence of
all-fail exposure, not proof of immunity. Noise robustness is tested only for outcome filtering at
two levels. Finally, the official-DAPO comparison uses native trainer/filter code but a custom
shaped-reward hook. DAPO itself uses accuracy-based filtering; we quantify a semantic risk of
repurposing other documented metrics under a composite reward.

\section{Practical rules}

\begin{enumerate}\itemsep2pt
\item When filtering outcome-degenerate groups under a composite reward, use an
      \textbf{outcome-semantic signal decoupled from the shaping term}. If no reliable binary
      outcome exists, define and test the intended degeneracy predicate explicitly.
\item Dashboard two leading indicators: \textbf{all-fail group fraction} (the dose) and
      \textbf{mean response length}; in our collapse runs, length moves before validation accuracy.
\item Before adding shaping, measure the baseline all-fail frequency. Interpret this exposure
      relative to the model--task pair rather than model scale alone.
\end{enumerate}

\section{Reproducibility}

The core GSM8K gate, launcher and no-GPU offline tests are in verl-recipe PR \#125
\citep{verlrecipe2026}
(artifact commit \code{61dd7b6}; verl commit \code{e52747a}); the earlier multi-turn environment is
PR \#121. The official-DAPO control used verl commit \code{ad2e3c2}. The companion artifact contains
paper source, raw logs with resolved configurations, summaries and the figure script; plotted points
are extracted from logs and tables cross-checked separately. The DAPO launcher was not archived, so
this control is auditable from logs and shared reward code but not one-command rerunnable.
Runtime:
$\approx$50 min per 40-step arm at 1.5B (RTX 5090) and $\approx$60 min at 7B (RTX PRO 6000 96\,GB,
actor 53\,GB + vLLM 29\,GB $=$ 82/96\,GB).

\section*{Acknowledgments}

Experiments ran on a single shared RTX 5090 workstation and a rented RTX PRO 6000 96\,GB instance.
This work builds entirely on the open-source verl / HybridFlow framework
\citep{sheng2024hybridflow} and its community \code{recipe/dapo} implementation, whose
configurability made the question in this paper askable in the first place.

\bibliographystyle{plainnat}
\bibliography{references}

\end{document}